\documentclass{article} 
\usepackage{iclr2015,times,amsmath,amssymb,dcolumn,subcaption,graphicx}
\usepackage{hyperref}
\usepackage{url}

\newcommand{\vect}[1]{\boldsymbol{#1}}
\newcolumntype{.}{D{.}{.}{-1}}

\title{Distributed Decision Trees}

\author{
Ozan \.Irsoy \\
Department of Computer Science\\
Cornell University\\
Ithaca, NY 14850, USA \\
\texttt{oirsoy@cs.cornell.edu} \\
\And
Ethem Alpayd{\i}n \\
Department of Computer Engineering \\
Bo{\u g}azi{\c c}i University\\
Bebek, {\.I}stanbul 34342 Turkey\\
\texttt{alpaydin@boun.edu.tr} \\
}

\iclrconference 

\begin{document}

\maketitle

\begin{abstract}
Recently proposed budding tree is a decision tree algorithm in which
every node is part internal node and part leaf. This allows representing
every decision tree in a continuous parameter space, and therefore a budding
tree can be jointly trained with backpropagation, like a neural network.
Even though this continuity
allows it to be used in hierarchical representation learning, the learned representations are local:
Activation makes a soft selection among all root-to-leaf paths in a tree. In this work
we extend the budding tree and propose the distributed tree where the children use different and independent splits and hence multiple paths
in a tree can be traversed at the same time. This ability to combine multiple paths gives the power
of a distributed representation, as in a traditional perceptron layer.
We show that distributed trees perform comparably or better than budding and traditional hard trees on
classification and regression tasks. 
\end{abstract}

\section{Introduction}

Decision tree is one of the most widely used models in supervised learning.
Consisting of internal decision nodes and terminal label nodes, it operates
by building a hierarchical decision function~\citep{cart, c45, rokachmaimon02}. For the input 
$\vect{x} = [1, x_1, \dots, x_d]$, this response function
can be defined recursively as follows:
\begin{align}
y_m(\vect{x}) =
\begin{cases}
\rho_m & \text{if } m \text{ is a leaf}\\
y_{ml}(\vect{x}) & \text{else if } g_m(\vect{x}) > 0\\
y_{mr}(\vect{x}) & \text{else if } g_m(\vect{x}) \leq 0\\
\end{cases}
\end{align}
If $m$ is a leaf node, for binary classification, $\rho_m \in [0,1]$
returns the probability of belonging to the positive class; for
regression, $\rho_m \in \mathbb{R}$ returns the scalar response value.
If $m$ is an internal node, the decision is forwarded to the left or
right subtree, depending on the outcome of the test $g_m(x)$.
$\vect{\rho}_m$ can be a vector as well, for tasks
requiring multidimensional outputs (e.g. multiclass classification,
vector regression).

There are many variants of the decision tree model based on how the
gating function $g_m(\cdot)$
is defined. Frequently,
\begin{align}
g_m(\vect{x}) &= x_{j(m)} - c_m 
\end{align}
is a gating function that compares one of the
input attributes to a threshold, and this is called the 
\emph{univariate tree}. The \emph{multivariate tree}~\citep{oblique, ldt} is a 
generalization in which the gating is linear,
\begin{align}
g_m(\vect{x}) &= \vect{w}^T \vect{x}
\end{align}
which allows arbitrary \emph{oblique} splits.
If we relax the linearity assumption on $g_m(\cdot)$, we have the
\emph{multivariate nonlinear tree}, e.g. in \citet{guo92}, $g_m(\cdot)$ is defined
as a multilayer perceptron. If the above assumptions are not fixed for the entire
tree but depend on the node $m$, we have the \emph{omnivariate tree}~\citep{omnivariate}.

Regardless of the gating and the leaf response functions, inducing optimal
decision tree is a difficult problem~\citep{rokachmaimon02}. Finding the smallest decision
tree that perfectly classifies a given set of input is NP-hard~\citep{hancocketal96}.
Thus, typically, decision trees are constructed greedily.

Essentially decision tree induction consists of two steps:

1) Growing the tree: Starting from the root node, at each node $m$, we search for the
best decision function $g_m(\vect{x})$ that splits the data that reaches the node $m$.
If the split provides an improvement in terms of a measure (e.g. entropy), we keep
the split, and recursively repeat the process for the children $ml$ and $mr$.
If the split does not provide any improvement, then $m$ is kept as a leaf and
$\rho_m$ is assigned accordingly.

2) Pruning the tree: Once the tree is grown, we can check if reducing the tree
complexity by replacing subtrees with leaf nodes
leads to an improvements on a separate development set. This is done to avoid
overfitting and improve generalization of the tree.

\section{Preliminaries}

\subsection{Soft Decision Trees}

Soft decision trees \citep{softtree,jordanjacobs94} build on top of multivariate linear trees by
softening the selection among the two children using a sigmoidal gating function,
instead of a hard selection. This allows the response function
to be continuous with respect to input $x$.

More formally, a soft decision tree models the response as the following recursive
definition:
\begin{align}
y_m(\vect{x}) =
\begin{cases}
\rho_m & \text{if } m \text{ is a leaf}\\
g_m(\vect{x}) y_{ml}(\vect{x}) + (1-g_m(\vect{x})) y_{mr}(\vect{x}) & \text{ otherwise}
\end{cases}
\end{align}
where $g_m(\vect{x}) = \sigma(\vect{w}_m^T \vect{x})$ with $\sigma(\cdot)$ being the standard
sigmoid (logistic) function.

Soft decision trees are trained incrementally in a similar fashion to hard trees.
Every node is recursively split until a stopping criterion is reached. Best split
is found by gradient descent on the splitting hyperplane of a parent and the response
values of the two children.

\subsection{Budding Trees}

Budding trees \citep{budtree} generalize soft trees further by softening the notion of
\emph{being a leaf} as well. Every node (called a \emph{bud} node) is part internal 
node and part leaf. For a node $m$ the degree of how much of
$m$ is a leaf is defined
by the \emph{leafness} parameter $\gamma_m \in [0, 1]$, which is a binary variable
for traditional decision trees ($\gamma_m \in \{0,1\}$). This allows the response
function to be continuous with respect to the parameters, including
the structure of a tree.

Formalization of the response function of a budding tree can thus be recursively defined as follows:
\begin{align}
y_m(\vect{x}) &= (1-\gamma_m)
 		\big[g_m(\vect{x}) y_{ml}(\vect{x}) + (1-g_m(\vect{x})) y_{mr}(\vect{x}) \big] + \gamma_m \rho_m
\label{eqn:bud}
\end{align}
The recursion ends when a node with $\gamma_m = 1$ is encountered.

Because of the continuity of the parameter space, budding trees are not limited to the
greedy incremental optimization scheme of decision tree induction. Exploiting the continuity,
we can employ backpropagation to compute gradients and
use continuous nonlinear optimization methods (e.g. stochastic gradient descent) to jointly
train an entire tree.

Note that every budding tree can be converted to a soft decision tree in which every internal
node has $\gamma_m = 0$ and every leaf has $\gamma_m = 1$. This is done by recursively distributing
the partial leaf contributions $\gamma \rho$ of internal nodes toward leaves. Therefore, one interpretation
for budding tree algorithm is that it is a way to train soft decision trees. This also means that
activation (selection) of a single path in a budding tree can be considered as an activation of a single
leaf node. This will simplify the remainder of the discussion.

\section{Distributed Trees}

\begin{figure}
\centering
\begin{subfigure}{0.49\textwidth}
\includegraphics{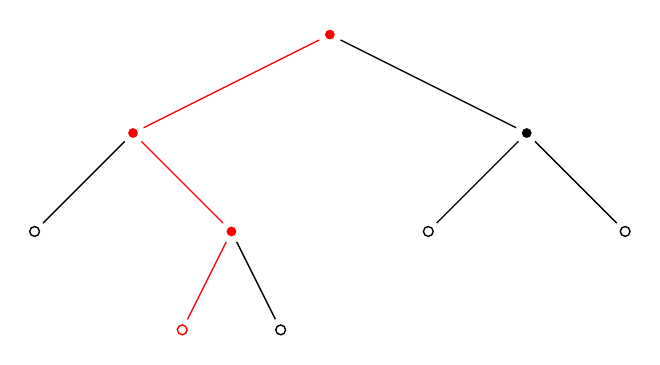}
\caption{}
\end{subfigure}
\begin{subfigure}{0.49\textwidth}
\includegraphics{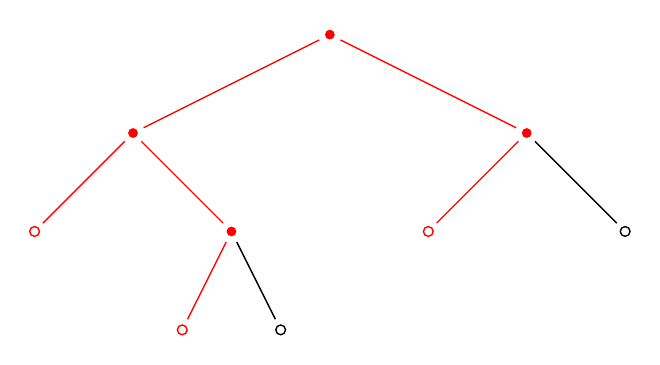}
\caption{}
\end{subfigure}
\caption{(a)~Operation of a budding tree which (softly) selects a single path across the tree.
(b)~Operation of a distributed tree where it can (softly) select multiple paths.}
\label{fig:budvsdist}
\end{figure}

Even though the budding tree provide a means for hierarchical representation learning, the resulting
representations are local. Because of its definition, the response of a budding tree is a convex
combination of all leaves. This provides a soft selection among the leaves that is akin to a
hierarchical softmax, and limits the representational power---If we use hard thresholds instead of
soft sigmoidal thresholds, it essentially selects one of the leaves. Thus, it partitions the
input space into as many parts as the number of leaves in the tree, which results in a representation
power that is linear in the number of nodes.

To overcome this limit, we extend the budding tree to construct a distributed tree where the two children of a node are selected by different and independent split conditions. Observe that the source of
locality in a budding tree comes from the convexity among leaves. Selecting one child more
means that it has to select the other less. We relax this constraint in the distributed tree
by untying the gating function that controls the paths for the two subtrees:
\begin{align}
y_m(\vect{x}) &= (1-\gamma_m)
 		\big[g_m(\vect{x}) y_{ml}(\vect{x}) + h_m(\vect{x}) y_{mr}(\vect{x}) \big] + \gamma_m \rho_m
\label{eqn:dist}
\end{align}
where $g_m(\vect{x}) = \sigma(\vect{w}_m^T)$ and $h_m(\vect{x})=\sigma(\vect{v}_m^T)$ are the conditions for the left and right children and $\vect{w}_m$, $\vect{v}_m$ respectively are the untied linear split parameters of the node $m$. Hence the conditions for left and right subtrees are independent---we get the budding (and traditional) tree if $h_m(\vect{x})\equiv 1-g_m(\vect{x})$.

With this definition, a tree no longer generates local representations, but distributed ones. The selection of one
child is independent of the selection of the other, and for an input, multiple paths can be traversed. Intuitively, a distributed tree becomes similar to a \emph{hierarchical sigmoid} layer as opposed to
a hierarchical softmax. In the case of hard thresholds instead of soft, more than one leaf node can be selected at the same time, implying any one of $2^{L}$ possibilities, where $L$ is the number of leaves (as opposed to exactly one of $L$ for the traditional local tree). This results in a representation power exponential in the number of nodes.

The distributed tree still retains the hierarchy that exists in traditional decision trees and budding trees.
Essentially, each node $m$ can still veto its entire subtree by not being activated. The activation of a node $m$ means that there is at least one leaf in the subtree that is relevant to this
particular input. With this interpretation, one can expect to see a hierarchy among the representations, relevant
features must be grouped together in subtrees.

\section{Experiments}

\begin{table}
\centering
\caption{Regression results}
\label{tab:regr}
\begin{tabular}{l|rrr|rrr}
\hline
&  \multicolumn{3}{c|}{MSE} & \multicolumn{3}{c}{Size} \\
\hline
& C4.5 & Bud & Dist & C4.5 & Bud & Dist\\
\hline
ABA & 54.13 & 41.61 & 41.79 & 44 & 35 & \textbf{24}\\ 
ADD & 24.42 &  4.68 & \textbf{ 4.56} & 327 & 35 & \textbf{27}\\ 
BOS & 34.21 & 21.85 & \textbf{18.28} & 19 & 19 & 33\\ 
CAL & 31.18 & 24.01 & \textbf{23.26} & 300 & 94 & \textbf{47}\\ 
COM & 3.61 & 1.98 & 1.97 & 110 & \textbf{19} & 29\\ 
CON & 26.89 & 15.62 & \textbf{15.04} & 101 & 38 & 43\\ 
8FH & 41.69 & 37.83 & 37.89 & 47 & \textbf{13} & 21\\ 
8FM & 6.89 & 5.07 & \textbf{5.02} & 164 & 17 & 15\\ 
8NH & 39.46 & \textbf{34.25} & 34.53 & 77 & \textbf{27} & 43\\ 
8NM & 6.69 & 3.67 & 3.61 & 272 & 37 & 37\\ 
\hline
\end{tabular}
\end{table}

\begin{table}
\centering
\caption{Binary classification results}
\label{tab:binary}
\begin{tabular}{l|rrrr|rrrr}
\hline
&  \multicolumn{4}{c|}{Accuracy} & \multicolumn{4}{c}{Size} \\
\hline
& C4.5 & LDT & Bud & Dist & C4.5 & LDT & Bud & Dist\\
\hline
BRE & 93.29 & 95.09 & 95.00 & 95.51 & 7 & \textbf{4} & 12 & 23\\ 
GER & 70.06 & \textbf{74.16} & 68.02 & 70.33 & \textbf{1} & 3 & 42 & 39\\ 
MAG & 82.52 & 83.08 & 86.39 & \textbf{86.64} & 53 & 38 & 122 & 40\\ 
MUS & 94.54 & 93.59 & 97.02 & \textbf{98.24} & 62 & 11 & 15 & 35\\ 
PIM & 72.14 & \textbf{76.89} & 67.20 & 72.26 & 8 & 5 & 68 & 35\\ 
POL & 69.47 & \textbf{77.45} & 72.57 & 75.33 & 34 & \textbf{3} & 61 & 97\\ 
RIN & 87.78 & 77.25 & 88.51 & \textbf{95.06} & 93 & \textbf{3} & 61 & 117\\ 
SAT & 84.58 & 83.30 & 86.87 & \textbf{87.91} & 25 & \textbf{9} & 38 & 41\\ 
SPA & 90.09 & 89.86 & 91.47 & \textbf{93.29} & 36 & \textbf{13} & 49 & 23\\ 
TWO & 82.96 & \textbf{98.00} & 96.74 & 97.64 & 163 & \textbf{3} & 29 & 25\\ 
\hline
\end{tabular}
\end{table}

\begin{table}
\centering
\caption{Multi-class classification results}
\label{tab:multi}
\begin{tabular}{l|rrrr|rrrr}
\hline
&  \multicolumn{4}{c|}{Accuracy} & \multicolumn{4}{c}{Size} \\
\hline
& C4.5 & LDT & Bud & Dist & C4.5 & LDT & Bud & Dist\\
\hline
BAL & 61.91 & 88.47 & 92.44 & \textbf{96.36} & 6 & \textbf{3} & 29 & 28\\ 
CMC & 50.00 & 46.65 & 53.23 & 52.87 & 25 & \textbf{3} & 28 & 51\\ 
DER & 94.00 & 93.92 & 93.60 & \textbf{95.84} & 16 & 11 & 11 & 20\\ 
ECO & 77.48 & 81.39 & 83.57 & 83.83 & 10 & 11 & 24 & 58\\ 
GLA & 56.62 & 53.38 & 53.78 & 55.41 & 21 & \textbf{9} & 21 & 55\\ 
OPT & 84.85 & 93.73 & 94.58 & \textbf{97.13} & 121 & \textbf{31} & 40 & 92\\ 
PAG & 96.72 & 94.66 & 96.52 & 96.63 & 24 & 29 & 37 & 27\\ 
PEN & 92.96 & 96.60 & 98.14 & \textbf{98.98} & 170 & 66 & \textbf{54} & 124\\ 
SEG & 94.48 & 91.96 & 95.64 & \textbf{96.97} & 42 & 33 & 33 & 76\\ 
YEA & 54.62 & 56.67 & 59.32 & 59.20 & 25 & 22 & 41 & 42\\ 
\hline
\end{tabular}
\end{table}

In this section, we report quantitative experimental results for 
regression, binary, and multi-class classification tasks.

We use ten regression (\emph{ABAlone}, \emph{ADD10}, \emph{BOSton}, \emph{CALifornia}, \emph{COMp}, 
\emph{CONcrete}, \emph{puma8FH}, \emph{puma8FM}, \emph{puma8NH}, \emph{puma8NM}), ten binary classification 
(\emph{BREast}, \emph{GERman}, \emph{MAGic}, \emph{MUSk2}, \emph{PIMa}, \emph{POLyadenylation},  \emph{RINgnorm}, 
\emph{SATellite47}, \emph{SPAmbase},  \emph{TWOnorm}) and
ten multiclass classification  (\emph{BALance}, \emph{CMC}, \emph{DERmatology}, \emph{ECOli},  
\emph{GLAss}, \emph{OPTdigits}, \emph{PAGeblock}, \emph{PENdigits}, \emph{SEGment}, \emph{YEAst})
data sets from the UCI repository~\citep{uci}, as in~\citet{budtree}. 
We compare distributed trees with budding trees and the C4.5 for regression and
classification tasks. Linear discriminant tree (LDT) which is a hard, multivariate tree \citep{ldt} is used as an additional baseline for the classification
tasks.

We adopt the following experimental methodology: We first separate one third of the data 
set as the test set over which we evaluate the final performance. With the remaining two thirds, 
we apply $5\times2$-fold cross validation. Hard trees (including the linear discriminant tree) use the validation set as a pruning set. 
Distributed and budding trees use the validation set to tune the learning rate and $\lambda$. 
Statistical significance is tested with the paired $t$-test for the performance measures, 
and the Wilcoxon Rank Sum test for the tree sizes, both with significance level $\alpha = 0.05$. 
The values reported are results on the test set not used for training or validation (model selection). 
Significantly best results are shown in boldface in the figures.

Table~\ref{tab:regr} shows the mean squared errors and the number of nodes 
of the C4.5, budding tree and distributed tree on the regression data sets.
Distributed trees perform significantly better on five data sets
(\emph{add10}, \emph{boston}, \emph{california}, \emph{concrete}, \emph{puma8fm}), whereas budding tree performs better
on one (\emph{puma8nh}), the remainder four being ties. In terms of tree sizes, both the distributed
tree and the budding tree has three wins (\emph{abalone}, \emph{add10}, \emph{california} and 
\emph{comp}, \emph{puma8fh}, \emph{puma8nh}, respectively), the remaining four are ties. Note that at the end of the training,
because of the stochasticity of the training, both distributed trees and budding trees have nodes
that are almost leaf (having $\gamma \approx 1$). These nodes can be pruned to get smaller trees
with negligible change in the overall response function.

Table~\ref{tab:binary} shows the percentage accuracy of C4.5, LDT, budding and distributed trees on 
binary classification data sets. In terms of accuracy, LDT has four wins (\emph{german}, \emph{pima}, 
\emph{polyadenylation}, \emph{twonorm}), distributed tree has five wins (\emph{magick}, \emph{musk2},
\emph{ringnorm}, \emph{satellite}, \emph{spambase}) and the remaining one (\emph{breast}) is a tie.
On its four win, LDT produces very small trees (and on three, it produces the smallest trees). This
suggests that with proper regularization, it would be possible to improve the performance of
budding and distributed trees. In terms of tree sizes, LDT has six wins (\emph{breast}, \emph{polyadenylation},
\emph{ringnorm}, \emph{satellite}, \emph{spambase}, \emph{twonorm})
and C4.5 has one (\emph{german}) with the remaining three ties.

Table~\ref{tab:multi} shows the percentage accuracy of C4.5, LDT, budding and distributed trees on
multiclass classification data sets. Distributed tree is significantly better on five data sets
(\emph{balance}, \emph{dermatology}, \emph{optdigits}, \emph{pendigits}, \emph{segment}), and the remaining
five are ties. In terms of tree sizes, again LDT produces smaller trees with four wins 
(\emph{balance}, \emph{cmc}, \emph{glass}, \emph{optdigits}), and budding tree has one win (\emph{pendigits}).

\section{Conclusions}

In this work we propose the model of the distributed trees which overcomes the locality of traditional trees and can learn distributed representations of the data. It does
this by allowing multiple root-to-leaf path (soft) selections over a tree structure, as opposed to (soft)
selection of a single path as done by budding and traditional trees. This increases their representational power from linear to exponential in the number of nodes. Quantitative evaluation on several data sets shows that this
increase is indeed helpful in terms of predictive performance. 

Even though the selection of left and right subtrees is independent in a distributed tree, it still preserves
the hierarchy in its tree structure as in traditional decision trees and budding trees. This is because
an activation close to zero for a node has the ability to veto its entire subtree, and a \emph{firing} node 
means that it believes that there is some relevant node in that particular subtree. This induces the tree to
gather relevant features together in subtrees, becoming finer and finer grained as it splits down further.

Previously we have proposed using decision trees autoencoders \citep{atnips} and we believe that distributed trees too can be considered as alternative to layers of perceptrons for deep learning in that they can learn hierarchical distributed representations of the input in its different levels. 
\bibliography{ref}
\bibliographystyle{iclr2015}

\end{document}